\documentclass[11pt]{article}

\PassOptionsToPackage{hyphens}{url}

\usepackage{acl}

\usepackage{times}
\usepackage{latexsym}
\usepackage[T1]{fontenc}
\usepackage[utf8]{inputenc}
\usepackage{microtype}
\usepackage{graphicx}
\graphicspath{{figures/}}

\usepackage{booktabs}
\usepackage{amsmath}
\usepackage{amssymb}
\usepackage{array}
\usepackage{xcolor}
\usepackage{colortbl}
\usepackage{framed}  

\definecolor{posG}{HTML}{1B7837}
\definecolor{negR}{HTML}{B2182B}

\title{Signal or Noise? A Benchmark Study of Agent Skills in Web Development}

\author{Ziyue Yang\thanks{\ These authors contributed equally.} \\
  Baidu, NLP \\
  Beijing, China \\
  \texttt{yangziyue01@baidu.com} \\\And
  Ding Fan\footnotemark[1] \\
  Baidu, NLP \\
  Beijing, China \\
  \texttt{dingfan06@baidu.com} \\}

\begin{document}
\maketitle

\begin{abstract}
Agent Skills are reusable procedural modules that are increasingly injected into coding-agent sessions to encode framework conventions, anti-patterns, and reusable tools. However, because each injected Skill expands the prompt of every query, an effective Skill benchmark must determine not only whether an agent can solve a task, but whether the Skill should have been injected at all. We introduce WebDev-Skills-Bench and use it for a controlled empirical study of 31 public WebDev Skills on 50 Web-Bench projects and 1,000 ordered tasks. The benchmark compares four matched conditions, including a length-matched irrelevant control and leave-one-out component ablations. To isolate Skill effects from prompt-length artifacts, we place only \texttt{SKILL.md} in the prompt while mounting auxiliary files into the agent workspace. Across four models, target Skill injection reduces mean Pass@2 by 1.3\% to 4.2\%, lowers task completion depth, and increases token cost by 72\% to 394\%, with gains in only 17\% to 36\% of Skill-project pairs. Length-matched controls reveal two failure modes: some models are length-distracted (an equally long irrelevant Skill reproduces most of the loss), while others are content-misled (prompt length is neutral, yet Skill content still lowers Pass@2 by 1.1\% to 1.4\%). Further analysis shows losses concentrate on easy early tasks, Skill rankings transfer weakly across models, and anti-pattern rules outperform example-heavy content within helpful Skills. These findings recast a matched Skill as a hypothesis about a particular (Skill, project, model) triple rather than a portable asset, reframing injection as a per-deployment routing decision and making length-matched controls and per-model audits a minimum standard for Agent-Skill evaluation.
\end{abstract}

\section{Introduction}
\label{sec:intro}

\textit{Agent Skills}~\cite{ref1,ref15} are reusable procedural modules, typically a Markdown file with optional auxiliary scripts and references, that are already deployed inside commercial coding agents and community tool stacks. Unlike a one-off instruction, a Skill acts as a persistent behavioral prior: the same content is injected into every prompt of a session to encode framework conventions, anti-patterns, and reusable tools. Vendor and community marketplaces now distribute thousands of such Skills~\cite{ref1,ref19,ref20,ref24,ref25}, and many agent configurations attach them at session start by default. This deployment pattern turns Skill evaluation into a benchmark problem. Every injected Skill enlarges the prompt of every query, and when the injected content does not help, the agent incurs token and latency overhead without a corresponding reliability gain, and in some cases a net loss. A WebDev Skill benchmark therefore needs to answer not only whether an agent can solve a task, but whether a matched Skill should have been injected at all.

Web development is a natural focus for this benchmark: JavaScript/TypeScript and HTML/CSS account for a large share of LLM coding traffic~\cite{ref2} and dominate developer activity more broadly~\cite{ref7,ref22}, and WebDev-targeted Skills are widely published. Yet the central measurement question remains open: does injecting a matched Skill improve agent performance on realistic WebDev tasks, or does it mostly enlarge the prompt without changing behavior? Existing evidence is conflicting and indirect. Native WebDev benchmarks such as Web-Bench~\cite{ref26} and ArtifactsBench~\cite{ref28} measure generation capability but never vary Skill injection, and cross-domain Skill benchmarks disagree on direction: SkillsBench reports a $+16.2$~pp gain over heterogeneous tasks~\cite{ref15}, whereas SWE-Skills-Bench finds 39 of 49 Skills yield zero pass-rate improvement on SWE-bench-style problems~\cite{ref9}. None isolates WebDev, and none separates Skill content from the prompt-length increase injection introduces.

We introduce \textbf{WebDev-Skills-Bench} and use it to conduct a controlled empirical study of 31 public WebDev Skills on Web-Bench's 50 projects and 1{,}000 ordered tasks. The benchmark is built around four matched conditions: \textbf{C0} (no Skill), \textbf{C1} (target Skill), \textbf{C2} (a length-matched irrelevant Skill that separates Skill content from prompt-length effects), and \textbf{C3} (a leave-one-out slice ablation that attributes a Skill's effect to its structural components: positive rules, anti-patterns, and example code). The core device is a \emph{workspace-aware injection protocol}: only \texttt{SKILL.md} enters the prompt, while auxiliary files (\texttt{references/}, \texttt{examples/}, \texttt{scripts/}) are mounted into the agent's filesystem, so prompt length is determined by \texttt{SKILL.md} alone and the length-matched control is well defined even for multi-file Skills.

The controlled study yields five findings. First, on all four models target injection produces a negative average $\Delta$Pass@2 (between $-1.3$ and $-4.2$~pp), lowers Task Completion Depth, and raises token cost by $72$ to $394$\%, while only $17$ to $36$\% of (Skill, project) pairs gain. Second, the C2 length-control splits this average into two mechanisms: Sonnet and Qwen are length-distracted (an equally long irrelevant Skill reproduces most of the loss), whereas GPT-5.1 and DeepSeek are content-misled (length is neutral but the content steers the model off-target). Third, the loss concentrates on easy early tasks rather than challenging ones---injection is most costly where the model already holds a strong prior. Fourth, per-pair Skill effects are nearly uncorrelated across models (cross-model Pearson $|r|\le0.12$ on $\Delta$Pass@2), which limits static cross-model recommendations. Fifth, decomposing a helpful Skill with C3 shows cheap anti-pattern rules carry its most reliable benefit, whereas example code helps weaker models but hurts the strongest, making example-heavy Skills a poor default.

We open-source the benchmark, the workspace-aware injection harness, and all per-(model, condition, pair) outputs.\footnote{\url{https://anonymous.4open.science/r/webdev-skills-bench-1C32/}} Together these recast Skill injection as a per-deployment routing decision---finding the beneficial minority before paying its injection cost---so length-matched controls and per-model audits should be a minimum standard for future Agent-Skill benchmarks.

\section{Related Work}
\label{sec:related}

\paragraph{Agent Skills as a prompting paradigm.}
Skills package procedural knowledge as prompt-readable Markdown plus optional workspace resources~\cite{ref1}. Unlike retrieval-augmented generation~\cite{ref14} or few-shot prompting~\cite{ref3}, a Skill supplies a persistent behavioral prior, since the same content is injected for every query in a session and encodes conventions, anti-patterns, and reusable tools. This makes Skills attractive for coding agents, whose failures often stem from framework-specific practice rather than syntax, but it also creates an evaluation challenge: a Skill can shift performance through its content, its length, or behavioral biases from framing the task as requiring an external module. Public Skill marketplaces~\cite{ref1,ref19,ref20,ref24,ref25} typically treat Skill quality as a property intrinsic to the Skill; we instead treat utility as a property of the (Skill, project, model) tuple.

\paragraph{Skill benchmarks.}
Recent Skill benchmarks disagree on whether Skills help on average. SWE-Skills-Bench~\cite{ref9} reports 39 of 49 Skills with zero pass-rate improvement on SWE-bench-style tasks, while SkillsBench~\cite{ref15} reports a $+16.2$~pp gain over heterogeneous domains. Neither isolates WebDev, and neither uses a length-matched irrelevant control to separate content from prompt-length effects; we hold the Web-Bench task harness fixed and vary only the Skill condition.

\paragraph{WebDev benchmarks.}
A growing body of work evaluates LLMs on web development with the prompt held fixed. These span functional code generation (Web-Bench~\cite{ref26}, ArtifactsBench~\cite{ref28}, WebApp1K~\cite{ref5}, WebGen-Bench~\cite{ref10}), design- or screenshot-to-code (Design2Code~\cite{ref21}, WebSight~\cite{ref13}, WebCode2M~\cite{ref8}, Web2Code~\cite{ref27}), and autonomous web agents (WebArena~\cite{ref29}, Mind2Web~\cite{ref6}). All measure capability on a fixed prompt; isolating the marginal effect of a matched Skill, as we do, is orthogonal to these efforts.

\section{Benchmark Design}
\label{sec:design}

\paragraph{Task corpus and Skill suite.}
WebDev-Skills-Bench is a pre-deployment Skill benchmark: it uses reproducible project tasks as a proxy for production WebDev work and measures the marginal effect of Skill injection rather than absolute model capability. We use Web-Bench~\cite{ref26} because, among recent WebDev benchmarks, it is not yet saturated by frontier models, relies on deterministic Playwright tests rather than LLM-as-judge scoring, and uses sequentially dependent tasks that expose long-horizon resilience. Web-Bench comprises 50 projects across 11 stack categories (React/Vue/Angular/Svelte front-ends, Express/Fastify back-ends, ORM/DB, CSS, Canvas/SVG/Three.js, bundlers, and DOM apps), with 20 ordered tasks per project, for 1{,}000 tasks in total.

All 31 Skills are third-party content drawn from prominent public repositories~\cite{ref1,ref19,ref20,ref24}; we authored none of them. Selection followed six principles: stack relevance, non-leakage, authoritative provenance, best-effort self-containedness, structural decomposability, and length coverage (the suite's \texttt{SKILL.md} files---the only text injected into the prompt---span roughly 1.2K--22K characters, so the byte-matched C2 control has comparable-length substitutes across the range). The Skill manifest and the full $31\times50$ routing matrix are in the supplementary material.

\paragraph{Routing.}
Indiscriminate Skill injection would mostly produce toolchain mismatches and dilute the utility signal. Two annotators (both authors, with professional WebDev experience) independently judged each of the $1{,}550$ (Skill, project) pairs as \emph{core} (the Skill's declared frameworks, libraries, or domain directly cover the project's primary stack) or \emph{skip} (no meaningful overlap), at roughly 5 minutes per pair---about 129 hours per annotator ($\approx$~258 person-hours in total). They agreed on $1{,}495$ pairs (raw agreement $96.5\%$, Cohen's $\kappa \approx 0.74$); the $55$ disagreements were resolved by discussion to consensus, with any unreconcilable pair assigned \emph{skip} to keep routing conservative. This yields \textbf{117 core pairs that cover all 50 projects}.

\paragraph{Workspace-aware injection.}
Many Skills ship auxiliary assets alongside \texttt{SKILL.md}; concatenating them into the prompt would inflate length unequally and break the length-matched control. We therefore inject only \texttt{SKILL.md} and mount the auxiliary directories into the agent's filesystem under \texttt{.skills/<skill-id>/}, so prompt length depends on \texttt{SKILL.md} alone and the byte-matched C2 control is tractable for any Skill.

\paragraph{Conditions.}
We hold the project workspace, execution harness, task order, decoding settings, and Playwright test suite fixed across conditions; the only intended variable is the prompt-level \texttt{SKILL.md} content. \textbf{C0} (native baseline) injects no Skill content and establishes intrinsic project difficulty, the zero-point for all $\Delta P_k = P_k(C_x) - P_k(C0)$. \textbf{C1} (target Skill) injects the core-matched Skill's \texttt{SKILL.md}; the pairwise difference $\Delta\mathrm{Total} = C1 - C0$ is the gross utility. \textbf{C2} (length-matched irrelevant Skill) replaces the target \texttt{SKILL.md} with a \emph{skip}-tier Skill of approximately equal byte length ($\pm 5\%$), which separates the target-content effect from the effect of adding a length-matched Skill block and gives $\Delta\mathrm{Length} = C2 - C0$ and $\Delta\mathrm{Content} = C1 - C2$. \textbf{C3} (leave-one-out slice ablation) removes one structural slice of the target \texttt{SKILL.md} at a time (positive rules $-R_p$, anti-patterns $-R_n$, or example code $-X$), attributing $\Delta\mathrm{Total}$ to its components on a focused subset of helpful pairs (\S\ref{sec:c3}). The C0--C2 contrasts measure \emph{whether} a Skill helps; the C3 contrast attributes \emph{which part} of it does (slice definitions and per-pair protocol in Appendix~\ref{app:c3}).

\begin{table*}[t]
\centering
\caption{Model-level C1$-$C0 effects on the 117 core pairs ($N{=}3$ per cell). Brackets give 95\% paired-bootstrap CIs. $\rho$ is relative total-token overhead.}
\label{tab:headline}
\small
\setlength{\tabcolsep}{5pt}
\begin{tabular}{lcccccc}
\toprule
Model & $N$ & $\Delta$P@1 (pp) & $\Delta$P@2 (pp) & $\Delta$TCD & Win/Tie/Loss (\%) & $\rho$ (tokens) \\
\midrule
Claude Sonnet 4   & 3 & $-2.8$ \,[$-4.6, -1.1$] & $-4.2$ \,[$-6.9, -1.9$] & $-0.85$ & $30/9/61$ & $+\phantom{0}72$\% \\
GPT-5.1           & 3 & $-1.6$ \,[$-3.3, -0.1$] & $-1.3$ \,[$-3.4, +0.5$] & $-0.26$ & $35/20/45$ & $+\phantom{0}74$\% \\
Qwen3 Coder 30B   & 3 & $-3.2$ \,[$-4.3, -2.1$] & $-2.3$ \,[$-3.4, -1.3$] & $-0.47$ & $17/35/48$ & $+\phantom{0}91$\% \\
DeepSeek V4 Flash & 3 & $-4.1$ \,[$-5.8, -2.2$] & $-2.0$ \,[$-3.8, -0.2$] & $-0.40$ & $36/16/48$ & $+394$\% \\
\bottomrule
\end{tabular}
\end{table*}

\begin{table*}[t]
\centering
\caption{$\Delta$Pass@2 (pp) by Web-Bench task difficulty ($N{=}3$); brackets are 95\% paired-bootstrap CIs. Bold marks a CI that excludes zero. A dash marks fewer than five pair-cells.}
\label{tab:chain-pos}
\small
\setlength{\tabcolsep}{6pt}
\begin{tabular}{lccc}
\toprule
Model & easy & moderate & challenging \\
\midrule
Sonnet 4          & \cellcolor{negR!20}$\mathbf{-5.3}$ \,[$-10.0, -0.7$] & $+3.3$ \,[$-2.7, +8.9$]   & $-2.6$ \,[$-11.5, +6.4$] \\
GPT-5.1           & \cellcolor{negR!20}$\mathbf{-4.0}$ \,[$-7.8, -0.4$]   & $-7.5$ \,[$-23.3, +7.5$]  & $+6.6$ \,[$-12.2, +22.7$] \\
Qwen3 Coder 30B   & \cellcolor{negR!20}$\mathbf{-10.7}$ \,[$-16.7, -5.4$] & \multicolumn{1}{c}{---}   & \multicolumn{1}{c}{---} \\
DeepSeek V4 Flash & \cellcolor{negR!20}$\mathbf{-7.3}$ \,[$-14.6, -0.5$]  & $+11.7$ \,[$+0.4, +24.3$] & $+0.8$ \,[$-10.0, +10.4$] \\
\bottomrule
\end{tabular}
\end{table*}

\paragraph{Models and metrics.}
We evaluate a four-model panel: Claude Sonnet~4 (\texttt{claude-sonnet-4-20250514}), GPT-5.1, DeepSeek-V4-flash, and Qwen3-Coder-30B-A3B. The panel spans the contrasts most likely to modulate Skill utility: closed frontier (Sonnet~4, GPT-5.1) versus open-weight (DeepSeek-V4-flash, Qwen3-Coder-30B), general-purpose versus coding-specialized (Qwen3-Coder), and larger versus smaller backends. All models use greedy decoding (\texttt{temperature}=0) with a 64\,k \texttt{maxTokens} budget, and each (model, condition, pair) cell is run for $N{=}3$ independent replicates so that within-condition variance is estimated uniformly across the panel. The workspace is reset (\texttt{git clean -fdx}) before every agent execution. All comparisons are paired at the (Skill, project) level. We report mean $\Delta$Pass@1 and $\Delta$Pass@2, mean $\Delta$Task Completion Depth (TCD, the longest consecutive Pass@2 prefix in the 20-task chain), and relative token overhead $\rho = (\mathrm{tokens}_{C_x} - \mathrm{tokens}_{C0})/\mathrm{tokens}_{C0}$, each with 95\% paired-bootstrap intervals over the pair set (1{,}000 resamples).

\section{Results}
\label{sec:results}

We organize the controlled empirical results around five claims about when Skill injection helps, hurts, and can be attributed.

\subsection{Average Skill injection does not justify its token cost}
\label{sec:headline}

Table~\ref{tab:headline} reports model-wise C1$-$C0 effects on the 117 core pairs. All four models show a negative mean $\Delta$Pass@2: Sonnet~4 $-4.2$~pp, Qwen $-2.3$, DeepSeek $-2.0$, and GPT-5.1 $-1.3$ (95\% CIs in Table~\ref{tab:headline}). Three of the four $\Delta$Pass@2 intervals exclude zero; GPT-5.1's marginally includes zero, but its $\Delta$Pass@1 ($-1.6$~pp $[-3.3, -0.1]$) does not. Task Completion Depth falls in parallel on every model ($-0.85$/$-0.47$/$-0.40$/$-0.26$ for Sonnet/Qwen/DeepSeek/GPT-5.1), so the negative average is not a pass-rate artifact. Token cost rises in parallel: $\rho$ is $+72$ to $+91$\% for three models and $+394$\% for DeepSeek (amplified by early C0 failures that shrink its denominator), confirming that C1 is cost-increasing.

The negative mean coexists with a positive-gain tail on every model: even Sonnet, whose mean effect is most negative, wins on $30\%$ of pairs, GPT-5.1 on $35\%$, DeepSeek on $36\%$, and Qwen on only $17\%$.

\subsection{The negative effect concentrates on easy tasks}
\label{sec:chain-position}

Web-Bench labels each of the 20 tasks per project as \textit{easy}, \textit{moderate}, or \textit{challenging}. A natural hypothesis is that Skills help most where they are most needed, that is, on challenging late-chain tasks. The evidence indicates the opposite. Table~\ref{tab:chain-pos} reports mean $\Delta$Pass@2 within each difficulty bucket, with per-pair-bucket CIs.

On every model the easy-task degradation is large enough that its CI excludes zero, from $-4.0$~pp (GPT-5.1) to $-10.7$~pp (Qwen). The moderate and challenging buckets are noisier---fewer pair-cells, with ceiling and floor effects---and show no consistent loss; DeepSeek even improves on its moderate tasks. Injection thus incurs its largest and most reliable losses on the tasks the model already handles correctly, so Skill utility should be reported by chain position rather than only as a project-level average.

A mechanism for this easy-task harm is \emph{retry lock-in}: on early tasks single-attempt mistakes are common but inexpensive, since Web-Bench's two-attempt budget recovers most of them when the model can vary simple structural choices (button text, class names, element nesting) between attempts. An injected Skill that fixes those choices in place, even when not technically wrong, converts recoverable first-attempt mistakes into chain-terminating failures and reduces self-repair flexibility on retry. Appendix~\ref{app:zustand} gives a concrete Sonnet $\times$ \emph{zustand} $\times$ \texttt{react-expert} trace.

\subsection{The negative average has two distinct mechanisms}
\label{sec:length-content}

\begin{table}[t]
\centering
\caption{Length/content decomposition (mean $\Delta$Pass@2 in pp, $N{=}3$). \emph{Survival} is the share of C1 wins that remain positive after the length control.}
\label{tab:length-content}
\small
\setlength{\tabcolsep}{3pt}
\begin{tabular}{lcccc}
\toprule
Model & $\Delta$Total & $\Delta$Length & $\Delta$Content & Survival \\
\midrule
Sonnet 4   & $-4.2$ & $-3.3$ & \cellcolor{negR!20}$-0.9$ & $60\%$ \\
Qwen        & $-2.3$ & $-3.5$ & \cellcolor{posG!20}$+1.2$ & $95\%$ \\
GPT-5.1     & $-1.3$ & $-0.2$ & \cellcolor{negR!20}$-1.1$ & $71\%$ \\
DeepSeek    & $-2.0$ & $-0.6$ & \cellcolor{negR!20}$-1.4$ & $64\%$ \\
\bottomrule
\end{tabular}
\end{table}

The length-matched C2 control decomposes the gross utility into a length artifact $\Delta\mathrm{Length} = C2 - C0$ and a content effect $\Delta\mathrm{Content} = C1 - C2$. Table~\ref{tab:length-content} shows that this decomposition partitions the panel into two groups.

\paragraph{Length distraction (Sonnet, Qwen).}
Sonnet's total loss ($-4.2$~pp) is largely accounted for by the length control ($\Delta\mathrm{Length} = -3.3$~pp, CI excludes zero), with a small content term that is not significant ($\Delta\mathrm{Content} = -0.9$~pp, CI includes zero); Qwen follows the same shape ($\Delta\mathrm{Length} = -3.5$, CI excludes zero; $\Delta\mathrm{Content} = +1.2$). For these two models the data fit attention dilution: a Skill-sized prompt block degrades performance, while the target content itself is not reliably worse than an equally long irrelevant block.

\paragraph{Content misalignment (GPT-5.1, DeepSeek).}
GPT-5.1 and DeepSeek show the opposite shape. Their length terms are near zero and their CIs include zero ($-0.2$ and $-0.6$~pp), while their content terms are negative ($-1.1$ and $-1.4$~pp). For these two models the prompt-length increase alone is essentially harmless, but injecting the specific content of the target Skill degrades performance. We read this as mechanism-level evidence: the same aggregate effect can arise from different causes across models, and the mitigation differs accordingly---prompt shortening versus content review.

\paragraph{Pair-level view.}
The \emph{Survival} column adds a complementary perspective: a majority of each model's C1 wins remain positive after the length control---$60\%$ for Sonnet, $71\%$ for GPT-5.1, $64\%$ for DeepSeek, and $95\%$ for Qwen---so even where the average effect is length-driven, most individual wins are content-positive rather than length artifacts. The decomposition thus describes model-level tendencies rather than certifying any individual pair; the per-pair content effect is what a benchmark should expose before routing.

\subsection{Cross-model contradictions caution against static Skill rankings}
\label{sec:cross-model}

\begin{table}[t]
\centering
\caption{Cross-model Pearson correlation of per-pair $\Delta$Pass@2 ($C1{-}C0$) over the 117 core pairs ($N{=}3$). All coefficients are near zero ($|r|\le0.12$; Spearman is similar, $|\rho_s|\le0.16$): a Skill's measured utility on one model barely predicts its utility on another.}
\label{tab:cross-model}
\small
\setlength{\tabcolsep}{5pt}
\begin{tabular}{lccc}
\toprule
 & DeepSeek & GPT-5.1 & Qwen \\
\midrule
GPT-5.1  & $-0.00$ &         &         \\
Qwen     & $+0.09$ & $+0.12$ &         \\
Sonnet 4 & $-0.07$ & $-0.06$ & $-0.08$ \\
\bottomrule
\end{tabular}
\end{table}

The decomposition in Table~\ref{tab:length-content} reports model-level statistics, but whether per-pair effects transfer across models is a separate, global property. Table~\ref{tab:cross-model} correlates each model's per-pair $\Delta$Pass@2 against every other's over the 117 core pairs. All six coefficients are near zero (Pearson $-0.08$ to $+0.12$, mean $\approx0.00$; Spearman similar), and disagreement is pervasive rather than confined to outliers: $74\%$ of pairs carry at least one positive and one negative model sign, while only $1\%$ gain on all four models and $4\%$ lose on all four.

This decorrelation is starkest in individual pairs: on \emph{lowdb} $\times$ \texttt{database-optimizer} (S14), Sonnet~4 gains $+33$~pp while DeepSeek and Qwen each lose $22$ and GPT-5.1 is unchanged---a $55$~pp swing on one core-tier pair that baseline difficulty does not explain (Appendix~\ref{app:cross-model}).

\paragraph{Static rankings transfer poorly across models.}
The near-zero cross-model correlation cautions against two common deployment shortcuts. First, ranking Skills by marketplace stars assumes one ranking transfers across backends; here the same content is a strong gainer for Sonnet and a clear liability for DeepSeek. Second, validating a Skill on a frontier model and deploying it on a cheaper one is unsafe, since the Sonnet signal ($+33$~pp on S14) gives no hint that the Skill should be withheld from DeepSeek. A practical router needs per-(Skill, project, model) traces, re-collected whenever any of the three vertices is upgraded.

\subsection{In helpful Skills, anti-patterns are the most cost-effective slice}
\label{sec:c3}

Conditions C0--C2 ask \emph{whether} a Skill helps; condition C3, a leave-one-out slice ablation, asks \emph{which part of it} does. We partition each \texttt{SKILL.md} into three removable slices (positive rules $R_p$, anti-patterns $R_n$, and example code $X$) and remove one at a time; a slice's \emph{contribution} is $C1-(\text{variant})$ on Pass@2. C3 runs on the positive tail---five (Skill, project) pairs with a robustly positive cross-model $C1{-}C0$ signal ($+3.3$ to $+6.7$~pp each)---so the complete Skill here raises Pass@2 by $+5.1$~pp, opposite in sign to the panel average (\S\ref{sec:headline}); the C3 numbers describe \emph{how a helpful Skill is built}, not how often Skills help. Pair selection, run protocol, statistical tests, and the selection caveat are in Appendix~\ref{app:c3}.

Table~\ref{tab:c3-slices} and Figure~\ref{fig:c3-slices} (Appendix~\ref{app:c3}) report per-slice contributions pooled over the 20 model$\times$project cells: anti-patterns ($R_n$) are the only slice with a directionally reliable task-level effect ($p{=}0.008$), positive rules ($R_p$) are neutral, and example code ($X$) is null on average ($-0.7$~pp). Yet $X$ is by far the costliest slice: removing it saves $34{,}482$ input tokens per run on average, about $22.7\%$ of the full \texttt{SKILL.md} budget---so cheap proscriptive content is the most cost-effective signal.

\begin{table}[t]
\centering
\caption{C3 slice contributions to Pass@2 (pp), $C1 - (\text{variant})$; cell-level means with paired-bootstrap CIs. ``All'' pools 20 cells; ``$-$Sonnet'' the 15 non-Sonnet cells. Positive = slice helps. Wilcoxon signed-rank and McNemar significance tests are reported in Appendix~\ref{app:c3}.}
\label{tab:c3-slices}
\small
\setlength{\tabcolsep}{4pt}
\resizebox{\columnwidth}{!}{%
\begin{tabular}{lcc}
\toprule
Slice & All ($N{=}20$) & $-$Sonnet ($N{=}15$) \\
\midrule
Whole Skill ($C1{-}C0$) & $+5.1$ \,[$1.0,9.3$] & $+6.1$ \,[$1.6,10.9$] \\
$R_p$ (positive rules)  & $+0.0$ \,[$-5.1,4.9$] & \cellcolor{posG!20}$+4.6$ \,[$1.1,8.4$] \\
$R_n$ (anti-patterns)   & $+3.1$ \,[$-3.3,9.8$] & $+6.1$ \,[$0.3,12.9$] \\
$X$ (examples)          & $-0.7$ \,[$-5.6,3.6$] & \cellcolor{posG!20}$+4.2$ \,[$2.0,6.8$] \\
\bottomrule
\end{tabular}%
}
\end{table}

The pooled example null, however, is an artifact of cancellation. Dropping Sonnet, every slice (including examples) turns positive, and the example effect becomes the strongest and most significant of all ($+4.2$~pp; Table~\ref{tab:c3-slices}, right). Figure~\ref{fig:c3-examples} (Appendix~\ref{app:c3}) shows the split: examples help DeepSeek ($+8.3$~pp) and Qwen ($+3.7$), are neutral for GPT-5.1 ($+0.7$), and clearly hurt Sonnet ($-15.3$). For the strongest model, in-skill examples act as a constraint that suppresses its own better priors---the retry-lock-in mechanism of \S\ref{sec:chain-position}. Example-heavy Skills should thus not be a default: they are expensive and model-dependent, whereas concise anti-pattern rules are most worth retaining.

\section{Benchmark Implications}
\label{sec:implications}

Taken together, these results make Skill injection a deployment decision rather than a configuration default. WebDev-Skills-Bench is intended as a pre-deployment audit for estimating whether a Skill's marginal gain survives its length cost and model-specific risks. The implications below translate this audit view into practical routing and reporting practices.

\paragraph{Benchmark injection as an opt-in decision.}
Unconditional injection is small-and-negative on every model and adds $72$ to $394$\% to token cost, so injecting a long Skill at session start, on average, reduces reliability while raising cost. Skills should be injected only when a pair-level signal (model, project, possibly task difficulty) crosses an empirical utility threshold.

\paragraph{Evaluate by chain position, not only by stack.}
Skill-induced degradation concentrates on easy initial tasks, where the model already produces the correct output. A reasonable heuristic is to skip the Skill on early tasks and inject one only once error rates rise---contradicting the common pattern of attaching Skills before any task begins.

\paragraph{Per-model curation is necessary.}
Per-pair effects are near-uncorrelated across models (Table~\ref{tab:cross-model}), so marketplaces that publish a single ranking are of limited use for multi-model deployment. A practical system needs per-model evaluation traces and a ranking conditioned on the model backend; a useful marketplace listing should report model-conditioned utility, the target stack, prompt length, and whether the gain survives a length-matched control.

\paragraph{A length-matched control should be the minimum bar.}
Without C2, we would have reported a uniform negative effect and missed that it arises from two mechanisms needing different mitigations. Future Agent-Skill benchmarks should adopt a length-matched control as a basic requirement; the workspace-aware protocol of \S\ref{sec:design} makes this tractable even for multi-file Skills.

\section{Conclusion}
\label{sec:conclusion}

We introduced WebDev-Skills-Bench, a controlled benchmark that asks not whether an agent can solve a task but whether a Skill should have been injected at all. Its contribution is methodological: a byte-matched control and a slice ablation turn a single negative average into a mechanistic account---a small positive tail, losses on easy tasks, and a length/content split across models that demands opposite mitigations. The claim is conditional: a Skill is a hypothesis about a particular (Skill, project, model) triple, not a portable asset, and capturing its value is a routing problem---finding the beneficial minority before paying its injection cost. We hope the controls that expose that minority become a default in Agent-Skill benchmarking.

\section*{Limitations}

Our study has several limitations. First, the seed spread is non-trivial: across the three Sonnet replicates, aggregate C0 and C1 Pass@2 vary by $4.4$ and $3.6$~pp, which is comparable to the headline effect size, so individual pair-level estimates should be read with corresponding caution even though the $N{=}3$ panel makes the model-level means stable. Second, our C2 measurements use $109$ unique length-matched runs across the $117$ pairs, because some C2 prompts are shared across same-project pairs; a cluster bootstrap and full per-pair C2 de-duplication are open follow-ups. Third, the routing is intentionally conservative, since C1 is evaluated only on the $117$ \emph{core}-tier pairs, so we cannot speak to what happens when Skills are deployed off-target. Fourth, the Skill set is drawn from high-visibility public repositories, and closed Skills from enterprise pipelines or fine-tuned skill-routers may produce different patterns. Fifth, WebDev-Skills-Bench is a pre-deployment benchmark rather than an online A/B test: Web-Bench projects approximate realistic WebDev work, but they do not include live user traffic, human developer interventions, or product-specific acceptance criteria. Finally, Web-Bench measures functional correctness via Playwright rather than visual fidelity or interactive UX, and we report token overhead rather than end-to-end latency, so Skills that primarily improve readability, accessibility, design quality, or developer review time may produce gains that our metrics do not capture.


\bibliography{custom}

@misc{ref1,
  author       = {{Anthropic}},
  title        = {Claude Agent Skills documentation},
  year         = 2025,
  howpublished = {\url{https://docs.claude.com/en/docs/agents-and-tools/agent-skills}},
  note         = {Accessed 2026-04}
}

@techreport{ref2,
  author       = {{Anthropic}},
  title        = {Anthropic Economic Index: {AI}'s Impact on Software Development},
  institution  = {Anthropic},
  year         = 2025,
  note         = {\url{https://www.anthropic.com/research/impact-software-development}}
}

@inproceedings{ref3,
  author       = {Tom B. Brown and Benjamin Mann and Nick Ryder and Melanie Subbiah and Jared Kaplan and Prafulla Dhariwal and Arvind Neelakantan and Pranav Shyam and Girish Sastry and Amanda Askell and Sandhini Agarwal and Ariel Herbert-Voss and Gretchen Krueger and Tom Henighan and Rewon Child and Aditya Ramesh and Daniel M. Ziegler and Jeffrey Wu and Clemens Winter and Christopher Hesse and Mark Chen and Eric Sigler and Mateusz Litwin and Scott Gray and Benjamin Chess and Jack Clark and Christopher Berner and Sam McCandlish and Alec Radford and Ilya Sutskever and Dario Amodei},
  title        = {Language Models are Few-Shot Learners},
  booktitle    = {Advances in Neural Information Processing Systems (NeurIPS)},
  year         = 2020,
  note         = {arXiv:2005.14165}
}

@article{ref5,
  author       = {Yi Cui},
  title        = {{WebApp1K}: A Practical Code-Generation Benchmark for Web App Development},
  journal      = {arXiv preprint arXiv:2408.00019},
  year         = 2024
}

@inproceedings{ref6,
  author       = {Xiang Deng and Yu Gu and Boyuan Zheng and Shijie Chen and Samuel Stevens and Boshi Wang and Huan Sun and Yu Su},
  title        = {{Mind2Web}: Towards a Generalist Agent for the Web},
  booktitle    = {Advances in Neural Information Processing Systems (NeurIPS) 36},
  year         = 2023,
  note         = {arXiv:2306.06070}
}

@misc{ref7,
  author       = {{GitHub}},
  title        = {Octoverse: A New Developer Joins {GitHub} Every Second as {AI} Leads {TypeScript} to \#1},
  year         = 2025,
  howpublished = {\url{https://github.blog/news-insights/octoverse/octoverse-a-new-developer-joins-github-every-second-as-ai-leads-typescript-to-1/}},
  note         = {Accessed 2026-04}
}

@article{ref8,
  author       = {Yi Gui and Zhen Li and Yao Wan and Yemin Shi and Hongyu Zhang and Yi Su and Bohua Chen and Dongping Chen and Siyuan Wu and Xing Zhou and Wenbin Jiang and Hai Jin and Xiangliang Zhang},
  title        = {{WebCode2M}: A Real-World Dataset for Code Generation from Webpage Designs},
  journal      = {arXiv preprint arXiv:2404.06369},
  year         = 2024
}

@article{ref9,
  author       = {Tingxu Han and Yi Zhang and Wei Song and Chunrong Fang and Zhenyu Chen and Youcheng Sun and Lijie Hu},
  title        = {{SWE-Skills-Bench}: Do Agent Skills Actually Help in Real-World Software Engineering?},
  journal      = {arXiv preprint arXiv:2603.15401},
  year         = 2026
}

@article{ref10,
  author       = {Zimu Lu and Yunqiao Yang and Houxing Ren and Haotian Hou and Han Xiao and Ke Wang and Weikang Shi and Aojun Zhou and Mingjie Zhan and Hongsheng Li},
  title        = {{WebGen-Bench}: Evaluating {LLMs} on Generating Interactive and Functional Websites from Scratch},
  journal      = {arXiv preprint arXiv:2505.03733},
  year         = 2025
}

@article{ref13,
  author       = {Hugo Lauren{\c c}on and L{\'e}o Tronchon and Victor Sanh},
  title        = {Unlocking the Conversion of Web Screenshots into {HTML} Code with the {WebSight} Dataset},
  journal      = {arXiv preprint arXiv:2403.09029},
  year         = 2024
}

@inproceedings{ref14,
  author       = {Patrick Lewis and Ethan Perez and Aleksandra Piktus and Fabio Petroni and Vladimir Karpukhin and Naman Goyal and Heinrich K{\"u}ttler and Mike Lewis and Wen-tau Yih and Tim Rockt{\"a}schel and Sebastian Riedel and Douwe Kiela},
  title        = {Retrieval-Augmented Generation for Knowledge-Intensive {NLP} Tasks},
  booktitle    = {Advances in Neural Information Processing Systems (NeurIPS) 33},
  year         = 2020,
  note         = {arXiv:2005.11401}
}

@article{ref15,
  author       = {Xiangyi Li and Wenbo Chen and Yimin Liu and Shenghan Zheng and Xiaokun Chen and Yifeng He and Yubo Li and Bingran You and Haotian Shen and Jiankai Sun and Shuyi Wang and Qunhong Zeng and Di Wang and Xuandong Zhao and Yuanli Wang and Roey Ben Chaim and Zonglin Di and Yipeng Gao and Junwei He and Yizhuo He and Liqiang Jing and Luyang Kong and Xin Lan and Jiachen Li and Songlin Li and Yijiang Li and Yueqian Lin and Xinyi Liu and Xuanqing Liu and Haoran Lyu and Ze Ma and Bowei Wang and Runhui Wang and Tianyu Wang and Wengao Ye and Yue Zhang and Hanwen Xing and Yiqi Xue and Steven Dillmann and Han-chung Lee},
  title        = {{SkillsBench}: Benchmarking How Well Agent Skills Work Across Diverse Tasks},
  journal      = {arXiv preprint arXiv:2602.12670},
  year         = 2026,
  note         = {Project page: \url{https://skillsbench.ai}}
}

@misc{ref19,
  author       = {{Mindrally}},
  title        = {Agent Skills Collection},
  year         = 2025,
  howpublished = {GitHub repository (MIT). \url{https://github.com/mindrally/skills}},
  note         = {Accessed 2026-04}
}

@misc{ref20,
  author       = {Addy Osmani},
  title        = {Web Quality Skills},
  year         = 2025,
  howpublished = {GitHub repository. \url{https://github.com/addyosmani/web-quality-skills}},
  note         = {Accessed 2026-04}
}

@article{ref21,
  author       = {Chenglei Si and Yanzhe Zhang and Zhengyuan Yang and Ruibo Liu and Diyi Yang},
  title        = {{Design2Code}: How Far Are We from Automating Front-End Engineering?},
  journal      = {arXiv preprint arXiv:2403.03163},
  year         = 2024
}

@misc{ref22,
  author       = {{Stack Overflow}},
  title        = {2025 Developer Survey Results},
  year         = 2025,
  howpublished = {\url{https://survey.stackoverflow.co/2025/}},
  note         = {Accessed 2026-04}
}

@misc{ref24,
  author       = {{Vercel Labs}},
  title        = {Agent Skills},
  year         = 2025,
  howpublished = {GitHub repository. \url{https://github.com/vercel-labs/agent-skills}},
  note         = {Accessed 2026-04}
}

@misc{ref25,
  author       = {{VoltAgent}},
  title        = {Awesome Agent Skills: A Curated Collection of 1000+ Agent Skills},
  year         = 2026,
  howpublished = {GitHub repository. \url{https://github.com/VoltAgent/awesome-agent-skills}},
  note         = {Accessed 2026-04}
}

@article{ref26,
  author       = {Kai Xu and YiWei Mao and XinYi Guan and ZiLong Feng},
  title        = {{Web-Bench}: A {LLM} Code Benchmark Based on Web Standards and Frameworks},
  journal      = {arXiv preprint arXiv:2505.07473},
  year         = 2025,
  note         = {ByteDance Research}
}

@article{ref27,
  author       = {Sukmin Yun and Haokun Lin and Rusiru Thushara and Mohammad Qazim Bhat and Yongxin Wang and Zutao Jiang and Mingkai Deng and Jinhong Wang and Tianhua Tao and Junbo Li and Haonan Li and Preslav Nakov and Timothy Baldwin and Zhengzhong Liu and Eric P. Xing and Xiaodan Liang and Zhiqiang Shen},
  title        = {{Web2Code}: A Large-Scale Webpage-to-Code Dataset and Evaluation Framework for Multimodal {LLMs}},
  journal      = {arXiv preprint arXiv:2406.20098},
  year         = 2024
}

@article{ref28,
  author       = {Chenchen Zhang and Yuhang Li and Can Xu and Jiaheng Liu and Ao Liu and Changzhi Zhou and Ken Deng and Dengpeng Wu and Guanhua Huang and Kejiao Li and Qi Yi and Ruibin Xiong and Shihui Hu and Yue Zhang and Yuhao Jiang and Zenan Xu and Yuanxing Zhang and Wiggin Zhou and Chayse Zhou and Fengzong Lian},
  title        = {{ArtifactsBench}: Bridging the Visual-Interactive Gap in {LLM} Code Generation Evaluation},
  journal      = {arXiv preprint arXiv:2507.04952},
  year         = 2025,
  note         = {Tencent Hunyuan}
}

@inproceedings{ref29,
  author       = {Shuyan Zhou and Frank F. Xu and Hao Zhu and Xuhui Zhou and Robert Lo and Abishek Sridhar and Xianyi Cheng and Tianyue Ou and Yonatan Bisk and Daniel Fried and Uri Alon and Graham Neubig},
  title        = {{WebArena}: A Realistic Web Environment for Building Autonomous Agents},
  booktitle    = {International Conference on Learning Representations (ICLR)},
  year         = 2024,
  note         = {arXiv:2307.13854}
}

\appendix

\section{Worked Example: Retry Lock-In on \emph{zustand} $\times$ \texttt{react-expert}}
\label{app:zustand}

This trace illustrates the easy-task retry lock-in of \S\ref{sec:chain-position}. \textit{Zustand} is a React state-management project, and \texttt{react-expert} (S07) is the most stack-aligned Skill routed to it. On the Sonnet C0 baseline the project reaches $55\%$ Pass@2 (mean over three seeds); injecting S07 lowers it to $40\%$ ($\Delta=-15$~pp), and the loss stems from a single seed whose chain terminates at task-4. Task-4 asks the model to create a \texttt{BlogForm} modal titled ``Create Blog'' with an ``Add Blog'' button in the header. On both C0 and C1 the first attempt emits a modal whose \texttt{<h2>Create Blog</h2>} heading and submit \texttt{<button>Create Blog</button>} collide under the same Playwright locator, and both fail with the identical strict-mode trace (Figure~\ref{fig:retry-lockin}). The conditions diverge on the retry: the C0 attempt relabels the submit button to \texttt{Submit}, so the locator resolves to a single element and the project continues to task-11, whereas the C1 attempt relabels it to \texttt{Create Blog Post}, in keeping with the Skill's prescription that button labels restate the operation; because the locator matches text as a substring, \texttt{Create Blog Post} still collides with the heading, strict mode fails again, and the chain terminates at task-4.

\begin{figure}[!h]
\small
\begin{framed}
\noindent\textbf{Both conditions, 1st attempt, fail with:}
\begin{verbatim}
strict mode violation:
getByText('Create Blog')
  resolved to 2 elements:
  1) <h2>Create Blog</h2>
  2) <button type="submit">
       Create Blog</button>
\end{verbatim}

\noindent\textbf{(a) C0 (no Skill) 2nd attempt: relabel, pass}
\begin{verbatim}
<button type="submit">
  Submit
</button>
\end{verbatim}

\noindent\textbf{(b) C1 (\texttt{react-expert}, S07) 2nd attempt: still collides, fail}
\begin{verbatim}
<button type="submit">
  Create Blog Post
</button>
\end{verbatim}
\end{framed}
\caption{Sonnet $\times$ \emph{zustand} $\times$ \texttt{react-expert}: the C0 retry relabels the colliding submit button to a structurally distinct string (a); the C1 retry, anchored on the Skill's naming convention, keeps the operation name and still matches the heading as a substring, so strict mode fails again (b).}
\label{fig:retry-lockin}
\end{figure}

\section{Content-Driven Win: \emph{pull-loading} $\times$ \texttt{js-dom-web-components}}
\label{app:pull-loading}

The model-level decomposition in \S\ref{sec:length-content} can conceal per-pair effects that exceed the panel mean by more than an order of magnitude. \emph{Pull-loading} is a vanilla-JavaScript pull-to-refresh widget routed to \texttt{js-dom-web-components} (S31), a Skill on DOM lifecycle and Web Component visibility patterns. On the Sonnet C0 baseline the project stays at $0\%$ Pass@2 across all three seeds, with the failure cascading from task-1, which asks the model to initialize the \texttt{content} and \texttt{noticeTxt} containers. Sonnet emits the requested DOM, but Playwright's \texttt{toBeVisible()} assertion fails because the empty wrapper has zero box dimensions; the C0 retry adds a defensive \texttt{display: block} style but never adds text content, so the second attempt fails for the same reason (Figure~\ref{fig:content-driven-win}a). Injecting S31 raises Pass@2 on the same pair to $18\%$, while the length-matched control returns to $0\%$, so $\Delta\mathrm{Content} = +18$~pp with $\Delta\mathrm{Length} = 0$. The mechanism is visible in the C1 second-attempt output (Figure~\ref{fig:content-driven-win}b): the model adds text inside the wrapper together with an explicit visibility style rule, directly applying S31's prescription that DOM nodes require both text content and explicit visibility properties to satisfy rendering assertions. Task-1 then passes and the project advances.

\begin{figure}[!h]
\small
\begin{framed}
\noindent\textbf{Task-1 Playwright assertion:} \verb|expect(noticeTxt).toBeVisible()|

\noindent\textbf{(a) C0 (no Skill), both attempts fail:}
\begin{verbatim}
<div id="content">
  <div id="noticeTxt"></div>
</div>
\end{verbatim}
\noindent Empty wrapper gives zero box dimensions, so \texttt{toBeVisible()} fails.

\noindent\textbf{(b) C1 (\texttt{js-dom-web-components}, S31), 2nd attempt: pass}
\begin{verbatim}
<div id="content">
  <div id="noticeTxt">
    Notice text content
  </div>
</div>

/* style.css */
#noticeTxt {
  display: block;
  visibility: visible;
}
\end{verbatim}
\end{framed}
\caption{Sonnet $\times$ \emph{pull-loading} $\times$ \texttt{js-dom-web-components}: an empty wrapper fails Playwright's visibility check (a); S31 adds text content and an explicit visibility rule, which satisfies the assertion (b). This single pair contributes a $+18$~pp content effect, far above Sonnet's $-0.9$~pp model-level average, which shows that the decomposition describes tendencies rather than per-pair guarantees.}
\label{fig:content-driven-win}
\end{figure}

\section{Cross-Model Sign Reversal: \emph{lowdb} $\times$ \texttt{database-optimizer}}
\label{app:cross-model}

The near-zero cross-model correlation of \S\ref{sec:cross-model} (Table~\ref{tab:cross-model}) is most vivid in individual pairs. Table~\ref{tab:cross-model-example} reports the single core-tier pair \emph{lowdb} routed to \texttt{database-optimizer} (S14), a mainstream back-end pair, as a concrete instance of the panel-wide decorrelation.

\begin{table}[!h]
\centering
\caption{Cross-model effects on \emph{lowdb} $\times$ \texttt{database-optimizer} (S14), an extreme, illustrative single pair ($N{=}3$). The same Skill content produces opposite-signed effects across the panel.}
\label{tab:cross-model-example}
\small
\setlength{\tabcolsep}{4pt}
\begin{tabular}{lccc}
\toprule
Model & C0 P@2 & C1 P@2 & $\Delta$P@2 (pp) \\
\midrule
Sonnet 4   & $33\%$ & $67\%$ & \cellcolor{posG!20}$+33$ \\
GPT-5.1    & $22\%$ & $20\%$ & $-2$ \\
DeepSeek   & $42\%$ & $20\%$ & \cellcolor{negR!20}$-22$ \\
Qwen        & $22\%$ & $\phantom{0}0\%$ & \cellcolor{negR!20}$-22$ \\
\bottomrule
\end{tabular}
\end{table}

The same Skill content produces opposite-signed effects: Sonnet~4 gains $+33$~pp (among the largest single-pair gains we observe), DeepSeek and Qwen each lose $22$, and GPT-5.1 is approximately unchanged. Baseline difficulty does not explain this: DeepSeek has the strongest C0 baseline ($42\%$) yet drops most, while Sonnet rises from $33\%$ to $67\%$. The $55$~pp gap between Sonnet and DeepSeek/Qwen on one core-tier pair is the extreme tail of the panel-wide pattern that a single deployment ranking cannot capture.

\section{C3 Slice Ablation: Protocol Detail}
\label{app:c3}

This appendix expands the C3 leave-one-out (LOO) protocol summarized in \S\ref{sec:c3}.

\paragraph{Slice definitions.}
Each \texttt{SKILL.md} is segmented into a meta/overview header (always retained), positive rules $R_p$ (prescriptive ``do'' conventions), anti-patterns $R_n$ (proscriptive ``don't'' rules), and example code $X$ (fenced code blocks demonstrating usage). Each ablation variant is the complete \texttt{SKILL.md} with exactly one slice removed ($-R_p$, $-R_n$, $-X$); the header is never removed so the prompt remains well-formed. A variant is marked N/A for any Skill that lacks the relevant slice, and only Skills with at least two removable slices are eligible.

\paragraph{Pair selection.}
Decomposition requires a gross effect to attribute. Pairs were screened by two rules on the cross-model main-experiment aggregates: the Skill must contain at least two cleanly removable slices (so there is something to decompose; this excludes, e.g., \emph{svelte}/S22 (rules only) and \texttt{backend-patterns}/S15 (examples only), whose \texttt{SKILL.md} reduces to a single removable slice after segmentation), and the pair must gain ($C1{>}C0$) on at least two of the four models (cross-model consistency rejects single-model noise). The five selected pairs are \texttt{javascript-pro} (S13) on \emph{vite} and \emph{svg-chart}, \texttt{bundler-config} (S29) on \emph{webpack}, \texttt{react-expert} (S07) on \emph{fastify-react}, and \texttt{database-optimizer} (S14) on \emph{sequelize}. On the final $N{=}3$ aggregates, \emph{vite}/S13, \emph{svg-chart}/S13, and \emph{webpack}/S29 gain on three of four models and \emph{fastify-react}/S07 and \emph{sequelize}/S14 gain on two; all five retain a positive cross-model mean gain ($+3.3$ to $+6.7$~pp). The released candidate table reports these aggregates. With only five pairs we make no per-domain claims and report contributions pooled and per model. This positive-tail selection is why the gross effect here ($+5.1$~pp) is opposite in sign to the panel average of \S\ref{sec:headline} and must not be read as evidence that Skills help on average.

\begin{figure}[!t]
\centering
\includegraphics[width=\columnwidth]{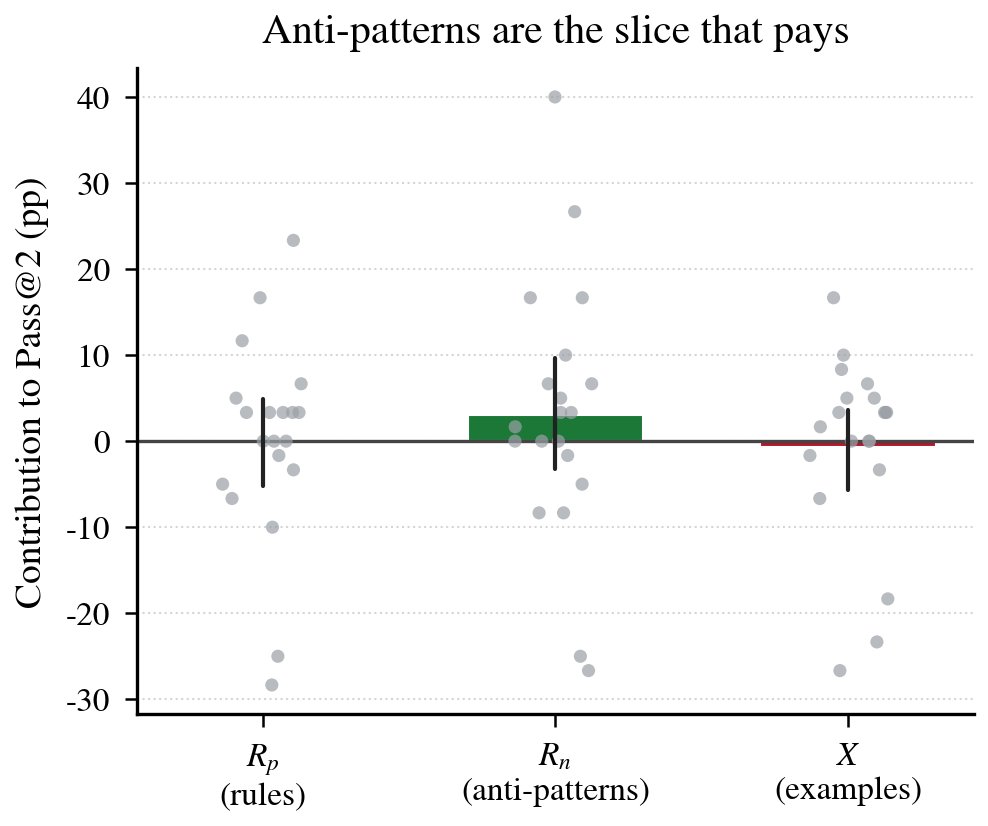}
\caption{Per-slice contribution to Pass@2, pooled over the 20 model$\times$project cells (bar = mean, whisker = paired-bootstrap CI; each gray dot is one cell). Bars are green when positive and red when negative, matching the shading of Table~\ref{tab:c3-slices}. Only anti-patterns ($R_n$) sit reliably above zero; positive rules and examples are null on average with wide per-cell spread.}
\label{fig:c3-slices}
\end{figure}

\begin{figure}[!t]
\centering
\includegraphics[width=\columnwidth]{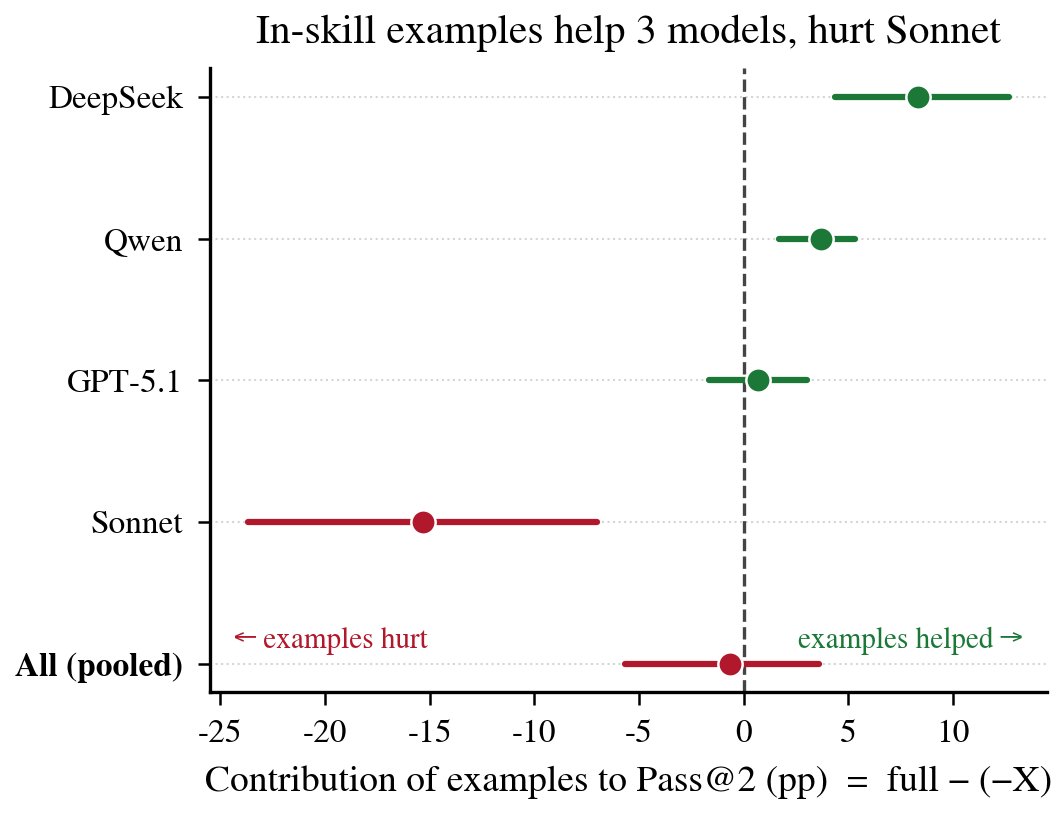}
\caption{Contribution of example code ($X$) to Pass@2 by model ($C1-(-X)$, cell mean with paired-bootstrap CI). Examples help DeepSeek, Qwen, and weakly GPT-5.1 but hurt Sonnet, so the pooled effect is near zero.}
\label{fig:c3-examples}
\end{figure}

\paragraph{Runs and statistics.}
Every variant is evaluated for all four models at $N{=}3$ seeds, anchored on the existing C0 and C1 runs, giving $5\times3\times3\times4=180$ ablation runs. The unit of inference is the cell (model~$\times$~project, $N{=}20$, seeds averaged). Per-slice contributions are tested with a Wilcoxon signed-rank test over cells with a $10{,}000$-resample paired-bootstrap CI; the task-level McNemar test (discordant counts $b,c$ over matched task attempts, missing tasks scored as failures) corroborates direction. For the pooled anti-pattern ($R_n$) contribution the McNemar discordants are $b{=}111$, $c{=}74$ ($p{=}0.008$); excluding Sonnet, the example-code ($X$) contribution of $+4.2$~pp has Wilcoxon $p{=}0.005$, the most significant slice effect in the panel. Figures~\ref{fig:c3-slices} and~\ref{fig:c3-examples} visualize the pooled per-slice contributions and the per-model split of the example-code effect. Per-pair contribution tables and the full per-(model, slice, seed) outputs are in the released artifacts.

\section{Released Artifacts}
\label{app:artifacts}

The analysis code, the routing for all conditions (C1 core pairs, C2 length-matched pairs, C3 candidate pairs), the 31 Skills with provenance, the C3 slice definitions, and the derived per-pair and per-task CSVs (C0--C3 values, pairwise $\Delta$s, per-slice C3 contributions, chain-position data, per-model rankings, and seed-variance tables) are released at \url{https://anonymous.4open.science/r/webdev-skills-bench-1C32/}. The repository includes the pipeline that reproduces the paper's tables and figures from these derived CSVs. The raw per-task evaluation reports and the base Web-Bench task harness are omitted for size and obtained from the upstream Web-Bench project, so the release fully reproduces the reported analyses but not a byte-for-byte rerun of every agent trajectory.

\end{document}